\documentclass{bmvc2k}
\usepackage{xinttools}
\usepackage{wrapfig}
\newcommand{\imwidth}{0.12\textwidth}

\title{Probabilistic Image Colorization}

\addauthor{Amelie Royer*}{aroyer@ist.ac.at}{1}
\addauthor{Alexander Kolesnikov*}{akolesnikov@ist.ac.at}{1}
\addauthor{Christoph H. Lampert}{chl@ist.ac.at}{1}

\addinstitution{
 IST Austria, Am Campus 1, \\ 3400 Klosterneuburg, Austria
}

\runninghead{Royer, Kolesnikov, Lampert}{Probabilistic Image Colorization}

\def\eg{{e.g}\bmvaOneDot} 
\def\ie{{i.e}\bmvaOneDot}

\newcommand{\METHODnospace}{PIC}
\newcommand{\METHOD}{\METHODnospace\xspace}
\newcommand{\embd}{g^w}
\newcommand{\pcnn}{f^{\theta}}

\begin{document}

\maketitle

\begin{abstract}
We develop a probabilistic technique for colorizing grayscale natural images.
In light of the intrinsic uncertainty of this task, the proposed probabilistic framework has numerous desirable properties.
In particular, our model is able to produce multiple plausible and vivid colorizations
for a given grayscale image and is one of the first colorization models to provide a
proper stochastic sampling scheme. 
Moreover, our training procedure is supported by a rigorous theoretical framework that
does not require any ad hoc heuristics and allows for efficient modeling and learning
of the joint pixel color distribution. 
%
We demonstrate strong quantitative and qualitative experimental results on the 
CIFAR-10 dataset and the challenging ILSVRC 2012 dataset.
\end{abstract}

\section{Introduction}
Colorization of natural grayscale images has recently been investigated in the deep learning community for its meaningful connection to classical vision tasks such as object recognition or semantic segmentation, as it requires high-level image understanding. In particular, its self-supervised nature (grayscale/color image pairs can be created automatically from readily available color images) allows for abundant and easy-to-collect training data; it has been shown that representations learned by colorization models are useful for -- and can be integrated in -- Computer Vision pipelines.
%

Previously proposed colorization models are able to capture the evident mappings abounding in the training data, \eg, blue sky, but often lack two main appealing properties: (i) \textit{diversity}, \ie being able to produce several plausible colorizations, as there is generally no unique solution, and (ii) \textit{color vibrancy} of the produced samples; the colorized images should display proper level of saturation and contrast like natural images, not look desaturated.

Most state-of-the-art colorization techniques do not in fact offer a proper sampling framework in the sense that they only model pixelwise color distributions rather than a joint distribution for colors of natural images. 
In contrast, our model relies on recent advances in autoregressive PixelCNN-type networks \cite{van2016conditional,kingma2016improving} for image modeling. Specifically, our architecture is composed of two networks. A deep feed-forward network maps the input grayscale image to an embedding, which encodes color information, much like current state-of-the-art colorization schemes. This embedding is fed to an autoregressive network, which predicts a proper distribution of the image chromaticity conditioned on the grayscale input.
Modeling the full multimodal joint distribution over color values offers a solution to the \textit{diversity} problem, as it provides us with a simple, computationally efficient, and yet powerful probabilistic framework for generating different plausible colorizations. Furthermore, the model likelihood can be used as a principled quantitative evaluation measure to assess the model performance.

As we discuss in the paper, the problem of color vibrancy is a consequence of not 
modeling pixel interactions and is hard to tackle in a principled way. 
In particular, \cite{zhang2016colorful} addresses it by (i) treating colorization as a classification task, avoiding the problem of using a regression objective which leads to unimodal, and thus, desaturated predictions, and (ii) introducing rebalancing weights to favor rare colors present in natural images and more difficult to predict.
In the experiments section, we show that our model generally produces vivid samples, without any \textit{ad hoc} modifications of the training procedure. 

In Section \ref{sec:methods} we introduce the theoretical framework to support the autoregressive component of our model, as well as our training and inference procedures. We report experimental 
results in Section~\ref{sec:exp}, including qualitative comparison to several recent baselines.

\section{Related work}
Automatic image colorization has been a goal of Image 
Processing and Computer Vision research since at least 
the 1980s, after movie studies started releasing 
re-colorized movies from the black-and-white era~\cite{f1972method}.
Because manually colorizing every frame of a movie is very tedious and expensive work, semi-automatic 
systems soon emerged, \eg based on the manual colorization 
of key frames followed by motion-based color
propagation~\cite{markle1988coloring}.
Subsequently, techniques that required less and less 
human interaction were developed, \eg, requiring only 
user scribbles~\cite{levin2004}, 
reference color images~\cite{Charpiat08ECCV,morimoto2009},
or scene labels~\cite{deshpande2015}. 

\begin{wrapfigure}{r}{0.33\textwidth}
  \begin{center}
    \includegraphics[width=0.32\textwidth]{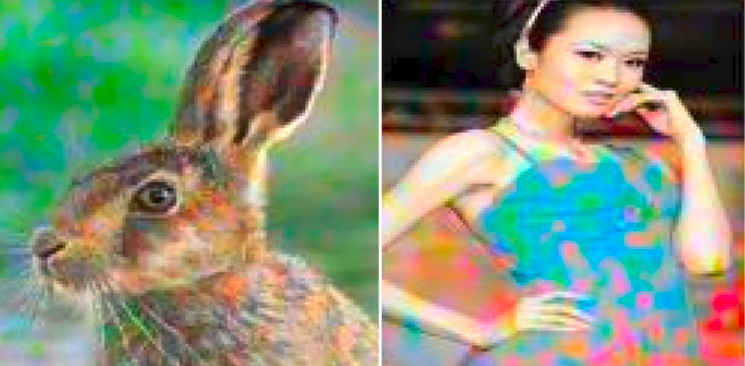}
  \end{center}
  \caption{Colorized samples from a feed-forward model.}\label{fig:noise}
\end{wrapfigure}

Succesful fully automatic approaches emerged only recently~\cite{IizukaSIGGRAPH2016,larsson2016learning,zhang2016colorful,isola2016,deshpande2016,DBLP:journals/corr/CaoZZY17}
based on deep architectures. 
A straight-forward approach is to train a convolutional feedforward 
model to independently predict a color value for each 
pixel~\cite{IizukaSIGGRAPH2016,larsson2016learning,zhang2016colorful}.   
However, these techniques do not model crucial interactions between
pixel colors of natural images, and thus,
probabilistic sampling yields high-frequency patterns of low perceptual
quality (see \hyperref[fig:noise]{Figure~\ref{fig:noise}}).
Predicting  the mode or expectation of the learned distribution instead results in grayish, and still often noisy colorizations (see, \eg, \hyperref[fig:compclose]{Figure \ref{fig:compclose}}).
%
%
Recent unpublished work~\cite{isola2016} proposes to 
train colorization model using generative adversarial
networks (GANs)~\cite{goodfellow2014generative}. 
GANs, however, are known to suffer from unstable training 
and lack of a consistent objective, which often 
prevents a quantitative comparison of models. 

A shared limitation of the models discussed above is their lack 
of \emph{diversity}. They can only produce one colored version 
from each grayscale image, despite the fact that are typically multiple
plausible colorizations.
\cite{DBLP:journals/corr/CaoZZY17} for instance addresses 
the problem in the framework of conditional GANs. To our knowledge,
the only work besides ours aiming at representing a fully probabilistic multi-modal joint distribution of pixel colors is~\cite{deshpande2016}.
It relies on the variational autoencoder framework~\cite{kingma2013auto}, which, however, tends to produce more blurry outputs than other image generating techniques. 
%
In contrast, the autoregressive~\cite{oord2016pixel,salimans2016pixel} network that we employ 
is able to produce crisp high-quality and diverse colorizations.



\section{Probabilistic Image Colorization}\label{sec:methods}

In this section we present our \textbf{P}robabilistic \textbf{I}mage \textbf{C}olorization model (\METHOD).
We first introduce the technical background, then formulate the proposed probabilistic model and conclude with parametrization and optimization details.

\subsection{Background}
Let $X$ be a natural image containing $n$ pixels, indexed in raster scan order: from top to bottom and from left to right; the value of the $i$-th pixel is denoted as $X_i$.
We assume that images are encoded in the LAB color space, which has three channels: the luminance channel (\textbf{L})
and the two chrominance channels (\textbf{a} and \textbf{b}). We denote by $X^L$ and $X^{ab}$ the projection of
$X$ to its luminance channel and chrominance channels respectively.
By convention a \textbf{Lab} triplet belongs to the range $[0; 100] \times [-127; 128] \times [-128; 127]$.
Consequently, each pixel in $X^{ab}$ can take $256 \times 256 = 65536$ possible values.

Our goal is to predict a probabilistic distribution of image colors from an input gray image (luminance channel), \ie we 
model the conditional distribution $p(X^{ab}|X^L)$ from a set of training images, $\mathcal{D}$. 
This is a challenging task, as $X^{ab}$ is a high dimensional object with a rich internal structure.

\subsection{Modeling the joint distribution of image colors}
To tackle the aforementioned task we rely on recent advances
in autoregressive probabilistic models~\cite{oord2016pixel,salimans2016pixel}.
The main insight is to use the chain rule in order to decompose the distribution of interest into 
elementary per-pixel conditional distributions; all of these distributions are modeled using a
shared deep convolutional neural network:
\begin{equation}
   p(X^{ab}|X^L) = \prod_{i=1}^n p(X_i^{ab}|X_{i-1}^{ab}, \dots, X_1^{ab};\ X^L).
   \label{eq:chain}
\end{equation}
Note, that \eqref{eq:chain} makes no assumptions on the modeled distribution.
It is only an application of the chain rule of probability theory.  
%
At training time, all variables in the factors are observed, so a model can be efficiently trained by learning all factors in parallel. 
At test time, we can draw a sample from the joint distribution using a pixel-level sequential procedure: we first sample $X_1^{ab}$ from $p(X_1^{ab}| X^L)$, then sample $X_i^{ab}$
from $p(X_i^{ab}| X_{i-1}^{ab}, \dots X_1^{ab}; X^L)$ for all $i$ in $\{ 2 \dots n \}$.

We denote the deep autoregressive neural network for modeling factors from \eqref{eq:chain} as $\pcnn$,
where $\theta$ is a vector of parameters.
The autoregressive network $\pcnn$ outputs a vector of normalized probabilities over
the set, $\mathcal{C}$, of all possible chrominance \textbf{(a, b)} pairs.
For brevity, we denote a predicted probability for the pixel value $X_i^{ab}$ as $\pcnn_i$.
To model the dependency on the observed grayscale image view $X^L$ we additionally introduce
a deep neural network $\embd(X^L)$, which produces a suitable embedding of $X^L$.
To summarize, formally, each factor in \eqref{eq:chain} has the following functional form:
\begin{equation}
   p(X_i^{ab}|X_{i-1}^{ab}, \dots, X_1^{ab};\ X^L) = \pcnn_i(X_{i-1}^{ab}, \dots, X_1^{ab};\ \embd(X^L))
   \label{eq:factor}
\end{equation}

Note, that the autoregressive network $\pcnn$ outputs a probability distribution over all color values in $\mathcal{C}$.
The standard way to encode such a distribution over discrete values is to parametrize $\pcnn$ to output a score for each of the possible color values in $\mathcal{C}$ and then apply the softmax operation to obtain a normalized distribution.
In our case, however, the output space is huge (65536 values per pixel), and the standard approach has crucial
shortcomings: it will result in a very slow convergence of the training procedure and will require a vast amount of data to generalize.
It is possible to alleviate this shortcoming by quantizing the colorspace at the expense of a slight drop in colorization accuracy and possible visible quantization artifacts. Furthermore, it still results in a large number of classes, typically a few hundreds, leading to slow convergence; additional heuristics, such as soft label encoding~\cite{zhang2016colorful}, are then required to speed up the training.

Instead, we approximate the distribution in \eqref{eq:factor} with a mixture of $10$ logistic distributions,
as described in \cite{salimans2016pixel}.
This requires $\pcnn$ to output the mixture weights as well as the first and second-order statistics of each mixture.
In practice, we need less than 100 output values per pixel to encode those, which is significantly fewer 
than for the standard discrete distribution representation.
This model is powerful enough to represent a multimodal discrete distribution over all values in $\mathcal{C}$.
Furthermore, since the representation is partially continuous, it can make use of the distance of the color values in the real space, resulting in faster convergence.

In the rest of the section section we give details on the architecture for $\embd$ and $\pcnn$ and on the optimization procedure.

\begin{figure}[!t]
\centering \includegraphics[width=0.95\textwidth]{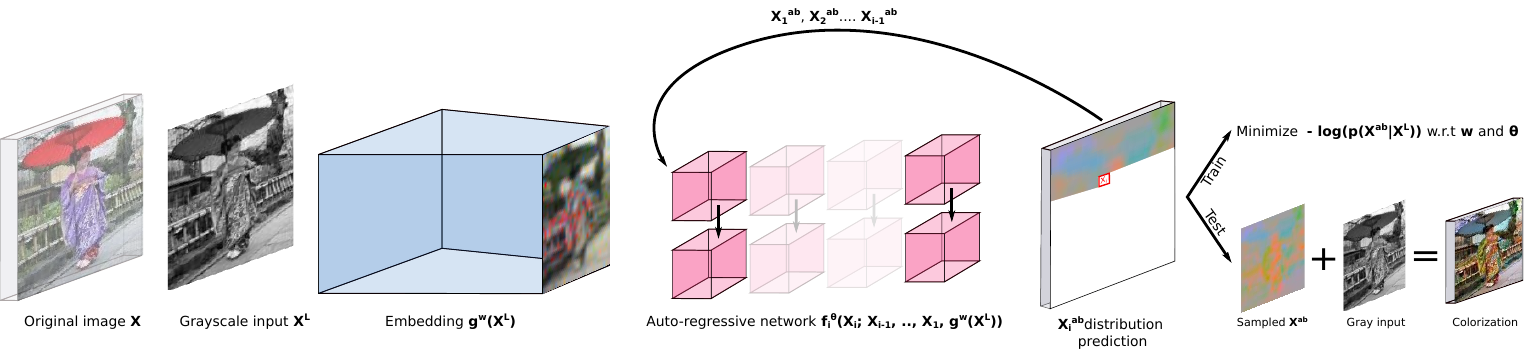}
\caption{\label{fig:architecture}High-level scheme for the proposed model.}
\end{figure}

\subsection{Model architecture and training procedure}
\label{sec:opti}

We present a high-level overview of our model in \hyperref[fig:architecture]{Figure~\ref{fig:architecture}}.
It has two major components: the embedding network $\embd$ and the autoregressive network $\pcnn$.
Intuitively, we expect that $\embd$, which only has access to the grayscale input,
produces an embedding encoding information about plausible image colors based on the semantics
available in the grayscale image.
The autoregressive network then makes use of this embedding to produce the final colorization,
while being able to model complex interactions between image pixels.

Our design choices for parametrizing networks $\embd$ and $\pcnn$ are motivated by \cite{salimans2016pixel},
as it reports state-of-the-art results for the challenging and related problem of natural image modeling.
In particular, we use \emph{gated residual blocks} as the main building component for the both networks.
Each residual block has 2 convolutions with 3x3 kernels, a skip connection~\cite{he2016deep}
and gating mechanism~\cite{van2016conditional,salimans2016pixel}.
Convolutions are preceded by concatenated~\cite{shang2016understanding} exponential linear
units~\cite{clevert2016fast} as non-linearities and parametrized as proposed in \cite{salimans2016weight}.
If specified, the first convolution of the residual block may have a dilated receptive field~\cite{YuKoltun2016};
we use dilation to increase the network's field-of-view without reducing its spatial resolution. 

The embedding network $\embd$ is a standard feed-forward deep convolutional neural network. It consist of gated residual
blocks and (strided) convolutions.
We give more precise details on the architecture in the experimental section.

For parametrizing $\pcnn$ we use the \emph{PixelCNN\nolinebreak[4]\hspace{-.01em}\raisebox{.2ex}{\small\bf ++}} architecture from~\cite{salimans2016pixel}.
On a high level, the network consists of two flows of residual blocks, where
the output of every convolution is properly shifted to achieve sequential
dependency: $X_i^{ab}$ depends only on $X_{i-1}^{ab}, \dots, X_1^{ab}$.
Conditioning on the external input, $X^L$, is achieved by biasing the output
of the first convolution of every residual block by the embedding $\embd(X^L)$.
We use no down- or up-sampling layers.
For more detailed explanation of this architecture see our implementation or \cite{salimans2016pixel}.

\paragraph{Spatial chromatic subsampling.}
It is known that the human visual system 
resolves color less precisely than 
luminance information~\cite{van1969spatiotemporal}.   
We exploit this fact by modeling the chrominance 
channels at a lower resolution than the input luminance.
%
%
This allows us to reduce computational and memory 
requirements without losing perceptual quality. 
Note that image compression schemes such as JPEG 
or previously proposed techniques for automatic 
colorization also make use of chromatic subsampling.

\paragraph{Optimization.} We train   
the parameters $\theta$ and $w$ by minimizing the 
negative log-likelihood of the chrominance channels in 
the training data:
\begin{equation}
        \arg\min\limits_{\theta, w} \sum_{X \in \mathcal{D}} -\log p(X^{ab}| X^L)
\end{equation}
We use the Adam optimizer~\cite{Kingma2014AdamAM} with 
an initial learning rate of 0.001, momentum  
of 0.95 and second momentum of 0.9995. 
We also apply Polyak parameter averaging~\cite{polyak1992acceleration}.


\begin{table}[t]
        \center
        \begin{tabular}[t]{|c|c|c|c|}
                \hline
                \multicolumn{4}{|c|}{CIFAR-10 embedding $\embd(X^L)$} \\
                \hline
                Operation & Res. & Width & D \\
                \hline
                Conv. 3x3/1  & 32 & 32 & -- \\
                Resid. block $\times$ 2 & 32 & 32 & -- \\
                Conv. 3x3/2  & 16 & 64 & -- \\
                Resid. block $\times$ 2& 16 & 64 & -- \\
                Conv. 3x3/1  & 16 & 128 & -- \\
                Resid. block $\times$ 2 & 16 & 128 & -- \\
                Conv. 3x3/1  & 16 & 256 & -- \\
                Resid. block $\times$ 3 & 16 & 256 & 2 \\
                Conv. 3x3/1  & 16 & 256 & -- \\
                \hline
        \end{tabular}
        \;
        \begin{tabular}[t]{|c|c|c|c|}
                \hline
                \multicolumn{4}{|c|}{ILSVRC 2012 embedding $\embd(X^L)$} \\
                \hline
                Operation & Res. & Width & D \\
                \hline
                Conv. 3x3/1  & 128 & 64 & -- \\
                Resid. block $\times$ 2& 128 & 64 & -- \\
                Conv. 3x3/2  & 64 & 128 & -- \\
                Resid. block $\times$ 2& 64 & 128 & -- \\
                Conv. 3x3/2  & 32 & 256 & -- \\
                Resid. block $\times$ 2& 32 & 256 & -- \\
                Conv. 3x3/1  & 32 & 512 & -- \\
                Resid. block $\times$ 3& 32 & 512 & 2 \\
                Conv. 3x3/1  & 32 & 512 & -- \\
                Resid. block $\times$ 3& 32 & 512 & 4 \\
                Conv. 3x3/1  & 32 & 512 & -- \\
                \hline
        \end{tabular}
        \caption{Architecture of $\embd$ for the CIFAR-10 and ILSVRC 2012 datasets.
                 The notation ``$\times\ k$'' in the \textbf{Operation} column means the corresponding
                 operation is repeated $k$ times.
                 \textbf{Res.} is the layer's spatial resolution,
                 \textbf{Width} is the number of  channels
                 and \textbf{D} is the dilation rate.}
        \label{table:embedding}
\end{table}

\section{Experiments}\label{sec:exp}

In this section we present quantitative and qualitative evaluation of the proposed
probabilistic image colorization (\METHOD) technique.
We evaluate our model on two challenging image datasets: CIFAR-10 and ImageNet ILSVRC 2012.
We also qualitatively compare our method to previously proposed colorization approaches and
perform additional studies to better understand various components of our model.
We will make our \textit{Tensorflow} implementation publicly available soon.

\subsection{CIFAR-10 experiments}

We first study the colorization abilities of our method on the CIFAR-10 dataset, which contains 50000 training images and 10000 test images of 32x32 pixels, categorized in 10 semantic classes.
We fix the architecture of the embedding network $\embd$
as specified in \hyperref[table:embedding]{Table~\ref{table:embedding} (left)}.
For the autoregressive network $\pcnn$ we use 4 residual blocks and 160 output channels
for every convolution.
We subsample the spatial chromatic resolution by a factor of 2, \ie
model the color channels on the resolution of 16x16.
We train the resulting model as explained in Section~\ref{sec:methods} 
with batch size of 64 images for 150 epochs.
The learning rate decays after every training iteration with constant multiplicative rate 0.99995.

\begin{figure}[t]
\centering
        \includegraphics[width=0.42\textwidth,trim={0 1.2cm 0 0},clip]{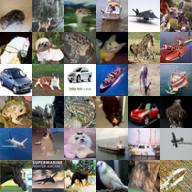}
        \quad
        \includegraphics[width=0.42\textwidth,trim={0 1.2cm 0 0},clip]{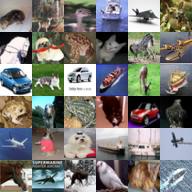}
        \caption{Colorized image samples from our model (left) and
                 the corresponding original CIFAR-10 images (right).
                 Images are selected randomly from the test set.}
        \label{fig:cifar-samples}
\end{figure}

In \hyperref[fig:cifar-samples]{Figure~\ref{fig:cifar-samples}} we visualize random test images colorized by \METHOD (left) and the corresponding
real CIFAR-10 color images (right).
We note that the samples produced by \METHOD appear to have natural colors and are hardly distinguishable from
the real ones.
This speaks in favour of our model being appropriate for modeling the color distribution of natural images.

We also report that \METHOD achieves a negative log-likelihood of 2.72, measured in bits-per-dimension.
Intuitively, this measure indicates the average amount of uncertainty in the image colors under the trained model.
This is a principled measure that can be used to perform model selection
and compare various probabilistic colorization techniques.

\subsection{ILSVRC 2012 experiments}

After preliminary experiments on the CIFAR-10 dataset
we now present experimental evaluation of \METHOD on the much more challenging ILSVRC 2012.
This dataset has 1.2 million high-resolution training images
spread over 1000 different semantic categories, and a hold-out set of 50000 validation images.
In our experiments we rescale all images to 128x128 pixels, which is enough to capture 
essential image details and remain a challenging scenario. 
Note, however, that in principle our method is applicable and scales to higher resolutions.

As ILSVRC images are of higher resolution and contain more details than CIFAR-10 images, 
we use a slightly bigger architecture for the embedding function $\embd$
as specified in \hyperref[table:embedding]{Table~\ref{table:embedding} (right)} and a chroma subsampling factor of 4, as in \cite{zhang2016colorful}.
The autoregressive component $\pcnn$ has 4 residual blocks and 160
channels for every convolution.

We run the optimization algorithm for 20 epochs using batches of 64 images,
with learning rate decaying multiplicatively after every iteration with the constant of 0.99999.

\begin{figure}[!t]
        \center
        \includegraphics[width=0.9\textwidth]{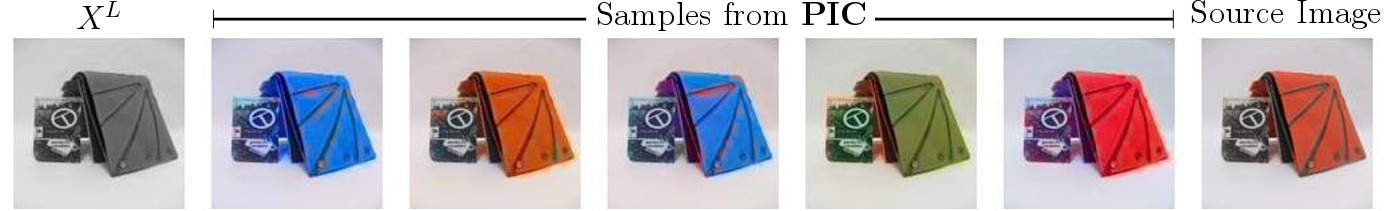}
        \includegraphics[width=0.9\textwidth]{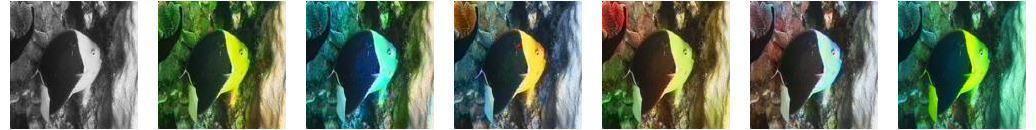}
        \includegraphics[width=0.9\textwidth]{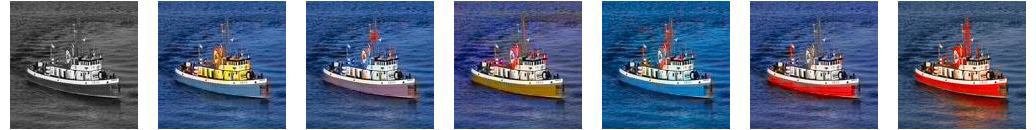}
        \includegraphics[width=0.9\textwidth]{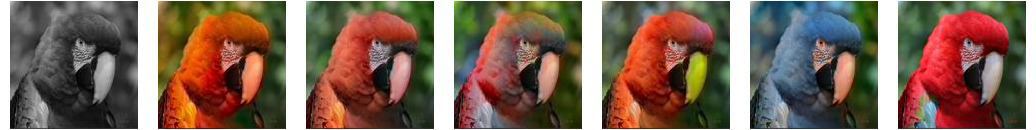}
        \vspace{0.5mm}
        \hrule
        \vspace{0.5mm}
        \includegraphics[width=0.9\textwidth]{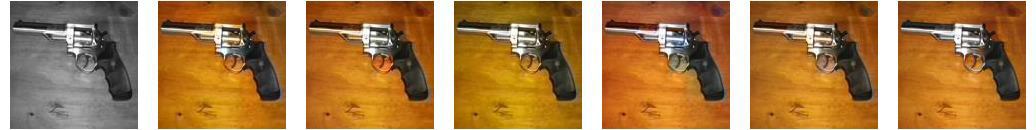}
        \includegraphics[width=0.9\textwidth]{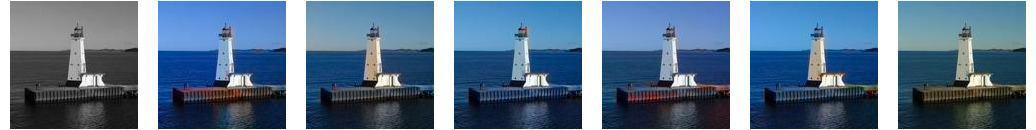}
        \includegraphics[width=0.9\textwidth]{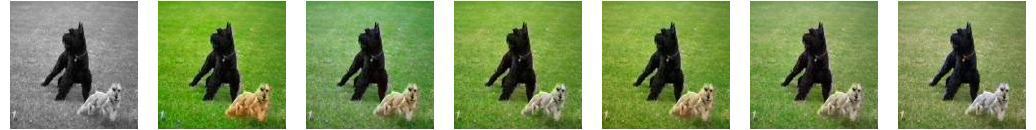}
        \caption{Colorized samples from our model illustrate its ability to produce
                 diverse (top) or consistent (bottom) samples depending whether the image semantics are ambiguous or not. 
                 }
        \vspace{-4mm}
        \label{fig:imagenet-samples}
\end{figure}

In \hyperref[fig:imagenet-samples]{Figure~\ref{fig:imagenet-samples}} we present successfully colorized images from the validation set.
These demonstrate that our model is capable of producing spatially coherent
and semantically plausible colors.
Moreover, as expected, in the case where the color is ambiguous, the produced samples often demonstrate
wide color diversity.
Nevertheless, if the color is mostly determined by the semantics of the object (grass or sky),
then \METHOD produces consistent colors.

\begin{figure}[!t]
        \center
        \includegraphics[width=0.9\textwidth]{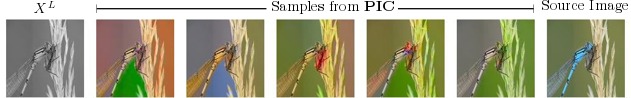}
        \includegraphics[width=0.9\textwidth]{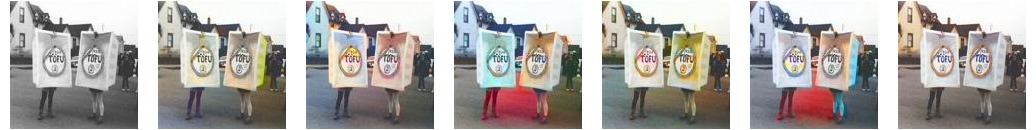}
        \caption{Illustration of failure cases: \METHOD may fail to reflect very long-range pixel interactions (top) and, \eg, assign  different colors to disconnected parts of an occluded object, or it may fail to understand semantics of complex scenes with unusual objects (bottom).}
        \label{fig:imagenet-failures}
\end{figure}

To provide further insight, we also highlight two failure cases in \hyperref[fig:imagenet-samples]{Figure~\ref{fig:imagenet-failures}}: 
First, \METHOD may not fully capture complex long-range pixel interactions interactions, \eg, if an object is split due to occlusion, the two parts may have different colors.
Second, for some complex images with unusual objects \METHOD may fail to understand 
semantics of the image and produce not visually plausible colors.

Our model achieves a negative log-likelihood of 2.51 bits-per-dimension.
Note that purely generative model from~\cite{oord2016pixel},
which is based on the similar but deeper architecture, reports a negative log-likehood of 
3.86 for the ILSVRC validation images modeled on the same resolution.
As our model has access to additional information (grayscale input),
it is not surprising that we achieve better likelihood; 
Nevertheless, this result confirms that \METHOD learns non-trivial colorization model
and strengthens our qualitative evaluation.

\subsection{Importance of the autoregressive component}

One of the main novelties of our model is the autoregressive component, $\pcnn$, 
which drastically increases the colorization performance by
modeling the joint distribution over all pixels.
In this section we perform an ablation study in order to investigate
the importance of the autoregressive component alone.
Note that without $\pcnn$, 
our model essentially becomes a standard feed-forward neural network, similar to recent colorization techniques~\cite{zhang2016colorful,larsson2016learning},
Specifically, we use \METHOD pretrained on the ILSVRC dataset,  
discard the autoregressive component $\pcnn$, and finetune
the remaining embedding network, $\embd$, for the task of image colorization.
At test time, we use maximum a posteriori (MAP) sampling from this model. Stochastic sampling from the output of $\embd$ would produce very noisy colorizations 
as the pixelwise predicted distributions are independent.
Alternatively, one could predict the mean color of the predicted distribution for each pixel, but that would produce mostly gray colors.

\begin{figure}[tb]
        \centering \includegraphics[width=0.485\textwidth]{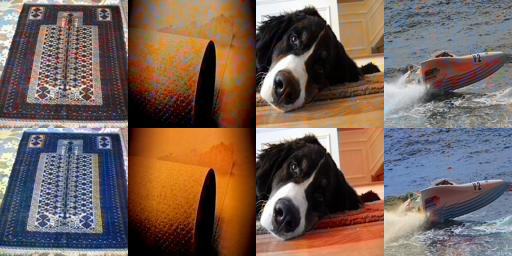}\;
        ~
                   \includegraphics[width=0.485\textwidth]{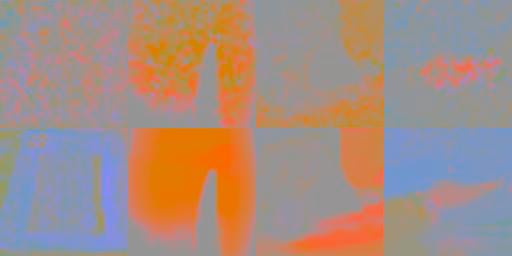}
        \caption{\label{fig:compclose} Comparison on ImageNet validation set between MAP samples from the
                                       embedding network $g_w$ \textit{(top)} and random samples from
                                       the autoregressive \METHOD model \textit{(bottom)}. Colored image (left) and predicted
                                       chrominances for fixed $\mathbf{L}=50$ (right)}
\end{figure}

From comparing the output samples of \METHOD and $\embd$, it appears the benefit brought by the autoregressive component is two-fold:
first, it explicitly models relationships between neighboring pixels, which leads to visually
smoother samples as can be seen in \hyperref[fig:compclose]{Figure \ref{fig:compclose}}. 
Second, the samples generated from \METHOD tend to display more saturated colors.
This is due to the fact that our model allows for proper probabilistic sampling
and, thus, can produce rare and globally consistent colors. 
We also verify that \METHOD produces more vivid colors by computing the average perceptual saturation~\cite{lubbe2010colours}.
Based on 1000 random image samples, the \METHOD model and $\embd$ have an average
saturation of 36.4\% and 32.7\%, respectively.

\subsection{Model selection and gating non-linearity}\label{gates}

\begin{figure}[h]
        \center
        \includegraphics[width=0.45\textwidth,trim={0 0 0 0},clip]{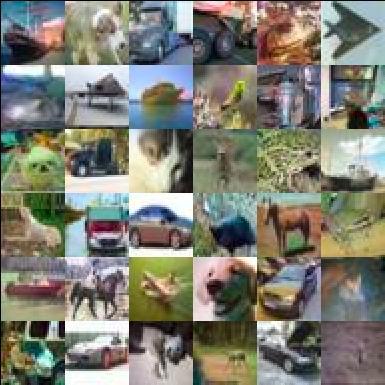}
        \quad
        \includegraphics[width=0.45\textwidth,trim={0 0 0 0},clip]{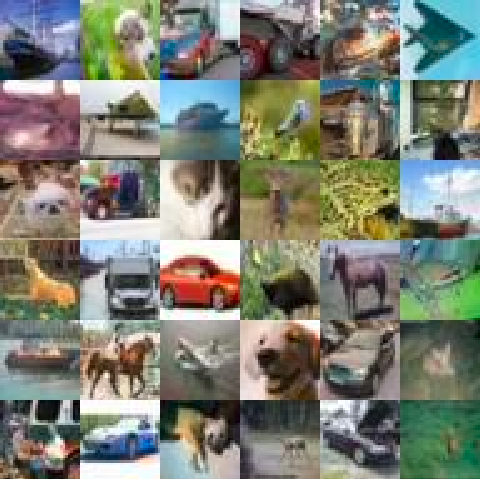}
        \caption{\label{fig:gating} Comparison of CIFAR-10 colorization samples obtained without gating
                                       \textit{(left)} and with gating \textit{(right)}.}
\end{figure}

Recently it was demonstrated that gating non-linearity is useful for the task of natural image modeling~\cite{oord2016pixel}.
Thus, we expect that gating should be also beneficial for colorization, which is closely related to image modeling.

One way to verify whether gating is useful is to perform qualitative sample analysis.
We illustrate samples from the \METHOD model, as well as the samples from the identical model without gating non-linearity
in Figure~\ref{fig:gating}.
We observe that even though the samples obtained with the gating mechanism appear to have higher visual quality,
\ie have slightly more saturated colors and better global consistency,
it is hard to make definitive conclusion.

Thus, we also perform quantitative analysis using the likelihood measure.
\METHOD with gating achieves
the negative log-likelihood of 2.72, while its counterpart without gating
achieves 2.78.
The quantitative evaluation is consistent with our preliminary qualitative evaluation.

We argue that the negative log-likelihood on the hold-out image set may be used
as a principled measure for model selection.
Importantly, our metric measures how well \emph{the joint distribution} of image
colors is explained by the model.
Unlike all previous metrics, which were used to evaluate image colorization performance,
our metric accounts for intrinsic uncertainty of the task and, at the same time,
for modeling complex interactions between pixels within one image.

\subsection{Qualitative comparison to baselines.}\label{sec:samples}

\begin{figure}[!t]
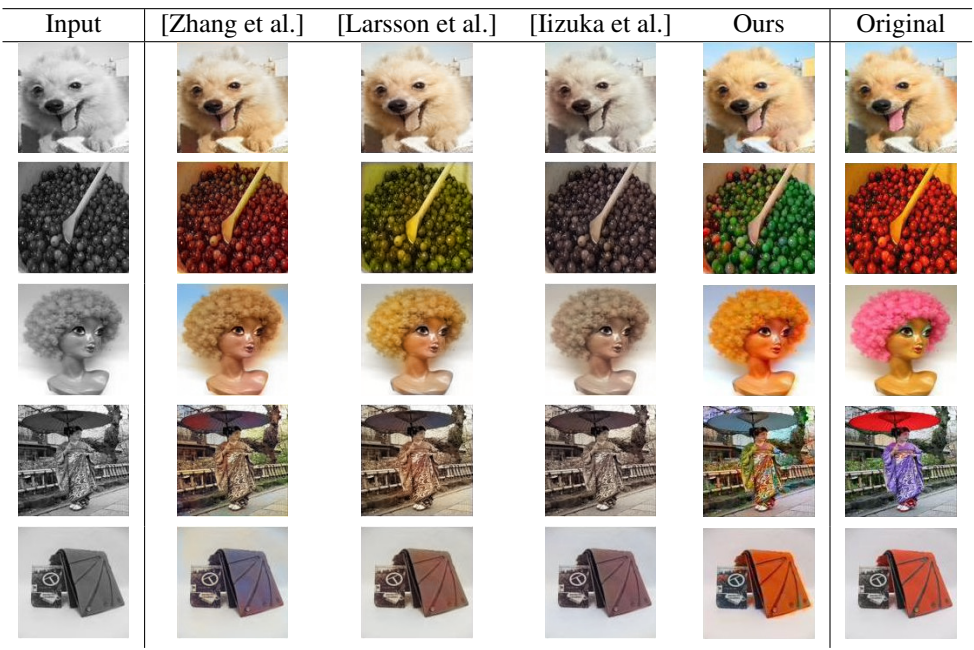

\begin{center}
\begin{tabular}{c|cccc|c}
\hline
Input & [Zhang et al.] & [Larsson et al.] & [Iizuka et al.] & Ours & Original\\
\hline
\xintFor* #1 in {3}
    \do
    {\includegraphics[width=\imwidth]{images/baselines/gray/#1.jpg}&
     \includegraphics[width=\imwidth]{images/baselines/colorful_colorization/#1.jpg}&
     \includegraphics[width=\imwidth]{images/baselines/Learning_Representations_for_Automatic_Colorizations/#1.png}&
     \includegraphics[width=\imwidth]{images/baselines/Neural_Network_based_Automatic_Image_Colorization/#1.jpg}&
     \includegraphics[width=\imwidth]{images/baselines/ours(selection)/#1.jpg}&\includegraphics[width=\imwidth]{images/baselines/original/#1.jpg}\\}
\xintFor* #1 in {1 2 3 5}
    \do
    {\includegraphics[width=\imwidth]{images/baselines/gray/#1_2.jpg}&
     \includegraphics[width=\imwidth]{images/baselines/colorful_colorization/#1_2.jpg}&
     \includegraphics[width=\imwidth]{images/baselines/Learning_Representations_for_Automatic_Colorizations/#1_2.png}&
     \includegraphics[width=\imwidth]{images/baselines/Neural_Network_based_Automatic_Image_Colorization/#1_2.jpg}&
     \includegraphics[width=\imwidth]{images/baselines/ours(selection)/#1_2.jpg}&\includegraphics[width=\imwidth]{images/baselines/original/#1_2.jpg}\\}
\end{tabular}
\end{center}
\caption{\label{fig:samples}Qualitative results from several recent automatic colorization methods compared to the original (right) and sample from our method (first to last column).}
\end{figure}

In \hyperref[fig:samples]{Figure \ref{fig:samples}} we present a few colorization results
on the ImageNet validation set for our model (one sample) as well as three recent colorization baselines.
\textbf{Zhang et al., 2016} \cite{zhang2016colorful} proposes a deep VGG architecture trained
on ImageNet for automatic colorization. The main innovation is that they treat colorization
as a classification rather than regression task, combined with class-rebalancing in the training
loss to favor rare colors and more vibrant samples.
\textbf{Larsson et al., 2016} \cite{larsson2016learning} is very similar to the first baseline, except
for a few architectural differences (\eg, use of hypercolumns) and heuristics.
\textbf{Iizuka et al., 2016} \cite{IizukaSIGGRAPH2016} proposes a non-probabilistic model with a regression objective. Their architecture is also more complex as as they use two distinct flows for local and global features. We also note that their model was trained on the MIT Places dataset, while ours and the two previous baselines use ImageNet.
We use the publicly available implementation for each baseline.

\begin{figure}[h!]
        \center
        \includegraphics[width=0.98\textwidth]{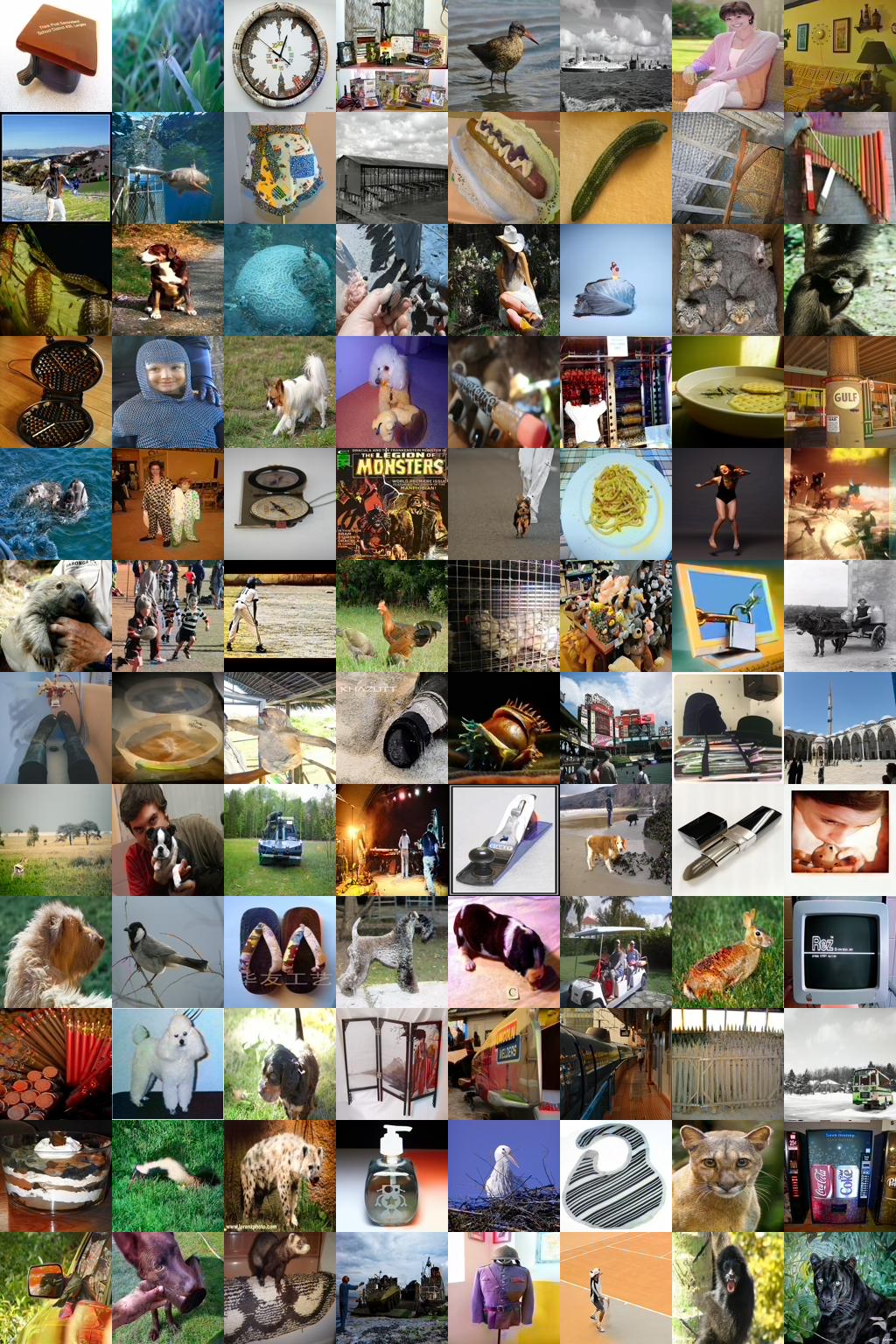}
        \caption{Random colorized images from the ILSVRC 2012
                 validation set \textit{(part2)}.}
        \vspace{-1mm}
        \label{fig:imagenet-random-2}
\end{figure}

In general, we observe that our model is highly competitive with other approaches
and tends to produce more saturated colors on average.
We will also include more samples from our model in the appendix, see Figures~\ref{fig:imagenet-random-1}~and~\ref{fig:imagenet-random-2}.

\section{Conclusion}

Deep feedforward networks achieve promising results on the task of colorizing natural gray images. The generated samples however often suffer of a lack of \textit{diversity} and \textit{color vibrancy}.
We tackle both aspects by modelling the full joint distribution of pixel color values using an autoregressive network conditioned on a learned embedding of the grayscale image.
The fully probabilistic nature of this framework provides us with a proper and straightforward  sampling mechanism, hence the ability to generate diverse samples from a given grayscale input. Furthermore, the data likelihood can be efficiently computed from the model and used as a quantitative evaluation metric.
We report quantitative and qualitative evaluations of our model, and show that colorizations sampled from our architecture often display vivid colors, indicating that the model captures well the underlying color distribution of natural images, without requiring any \textit{ad hoc} heuristics during training.

\bibliography{bmvc}

\clearpage
\section{Appendix}

\begin{figure}[h!]
        \center
	\includegraphics[width=0.98\textwidth,
                         trim={0 0 0 0},clip]{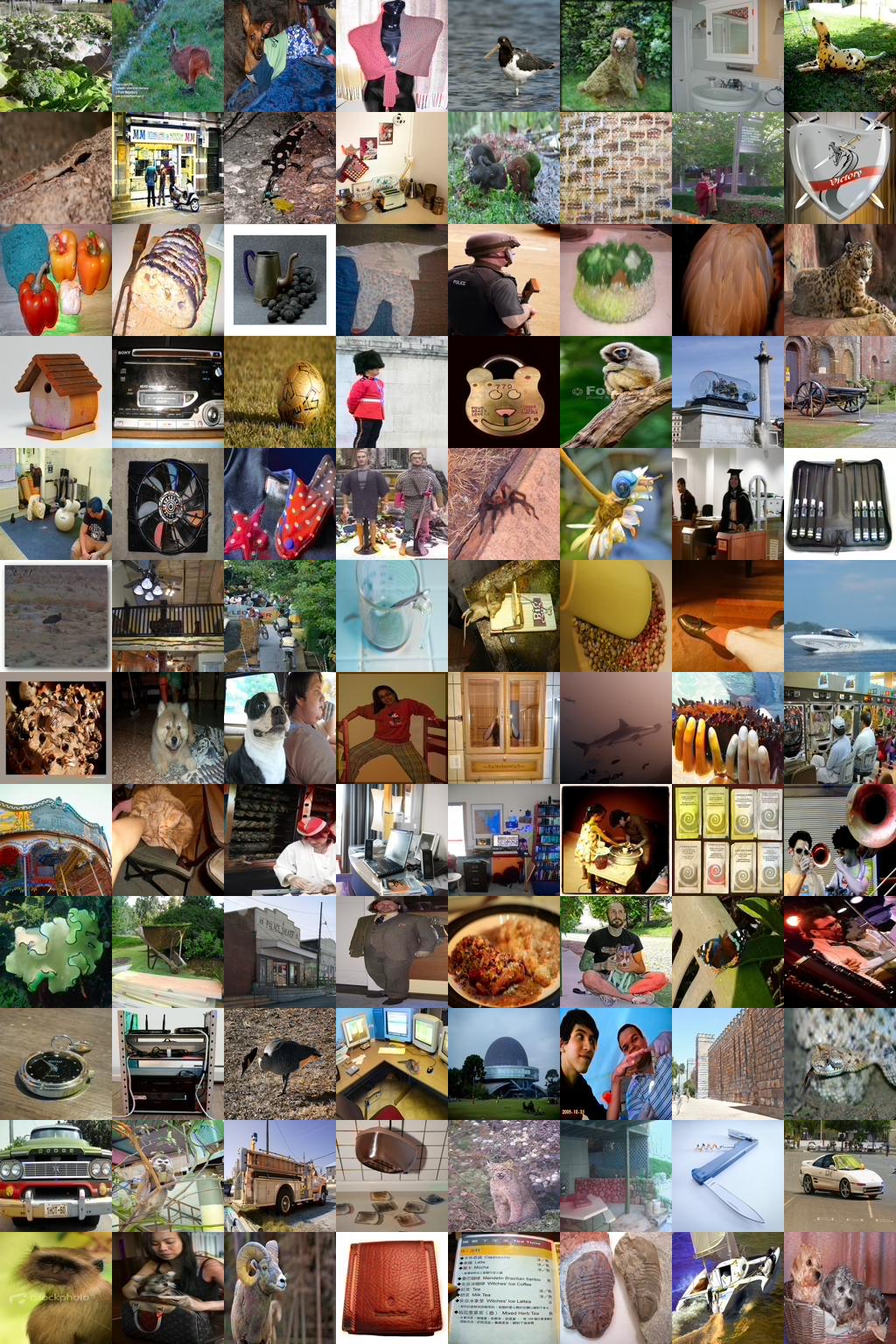}
        \caption{Random colorized images from the ILSVRC 2012 
                 validation set \textit{(part 1)}.}
        \vspace{-1mm}
        \label{fig:imagenet-random-1}
\end{figure}

\end{document}